\documentclass{bmvc2k}
\usepackage{amsfonts,amssymb}
\usepackage{mathtools}
\usepackage{multirow}
\usepackage{booktabs}
\usepackage{color}
\usepackage{adjustbox}
\usepackage{wrapfig}
\usepackage{graphicx}
\usepackage{subfigure}

\title{3D Object Tracking with Transformer}

\addauthor{Yubo Cui\textsuperscript{1,}}{ybcui21@stumail.neu.edu.cn
}{2}
\addauthor{Zheng Fang}{fangzheng@mail.neu.edu.cn}{1}
\addauthor{Jiayao Shan}{shanjiayao97@stumail.neu.edu.cn}{1}
\addauthor{Zuoxu Gu}{guzuoxu@stumail.neu.edu.cn}{1}
\addauthor{Sifan Zhou}{zhousifan@stumail.neu.edu.cn}{1}
\addinstitution{
	Northeastern University\\
	Shenyang, China
}
\addinstitution{
	Science and Technology on Near-Surface Detection Laboratory\\
	Wuxi, China
}
\runninghead{Cui ET AL.}{3D Object Tracking with Transformer}

\def\etal{\emph{et al}\bmvaOneDot}

\begin{document}

\maketitle

\begin{abstract}
	Feature fusion and similarity computation are two core problems in 3D object tracking, especially for object tracking using sparse and disordered point clouds. 
	Feature fusion could make similarity computing more efficient by including target object information. However, most existing LiDAR-based approaches directly use the extracted point cloud feature to compute similarity while ignoring the attention changes of object regions during tracking. In this paper, we propose a feature fusion network based on transformer architecture. Benefiting from the self-attention mechanism, the transformer encoder captures the inter-  and intra- relations among different regions of the point cloud. By using cross-attention, the transformer decoder fuses features and includes more target cues into the current point cloud feature to compute the region attentions, which makes the similarity computing more efficient. Based on this feature fusion network, we propose an end-to-end point cloud object tracking framework, a simple yet effective method for 3D object tracking using point clouds. Comprehensive experimental results on the KITTI dataset show that our method achieves new state-of-the-art performance. Code is available at: \url{https://github.com/3bobo/lttr}.
\end{abstract}

%-------------------------------------------------------------------------
\section{Introduction}
\label{sec:intro}
Recently, LiDAR-based 3D object tracking has been received more and more attention.
Benefiting from the development of visual tracking~\cite{SiamFC,GOTURN,SiamRPN,siamrpn++,ATOM}, most 3D tracking methods~\cite{SC3D, FSiamese, P2B} also use the Siamese-like tracking pipeline.
The pipeline first inputs template point clouds of the target object and search point clouds of the current frame to its top and bottom branches respectively, then fuses the two-branch features based on similarity. Finally, the fused features are used to localize the position of the object to be tracked.
However, compared with visual tracking, LiDAR-based tracking has more challenges due to the sparsity and disorder of the point clouds. 
For example, the point clouds will become much sparser with the increasing distance of the object, which hinders the feature extraction.
Meanwhile, the disorder of the point clouds also makes it hard to compute the similarity between the two branches.

Previous works use shape completion~\cite{SC3D}, image prior~\cite{FSiamese}, or feature augmentation~\cite{P2B} to deal with the above problems. Although they achieve better tracking performance, they usually ignore the attention changes in different regions of the object during tracking. However, the tracking method should pay more attention to regions with salient features when processing dense point cloud, while it should focus on regions with more points when processing sparse point cloud. Therefore, in the tracking process, different regions in the point cloud should have different attentions depending on the situation, even the same region also should have different attentions in different periods.

Inspired by~\cite{VIT, TNT}, in this work, we introduce transformer architecture~\cite{Transformer} into LiDAR-based 3D object tracking. First,  the point cloud is divided into several non-overlapping local regions. Then, based on the self-attention mechanism of the transformer encoder, the representation of each region is constructed by capturing the structural information of the local points, and the feature of the point cloud is reconstructed by considering the global relation among regions. Finally, in the decoding process, through propagating the template feature to the current search feature, the feature of the target object becomes more prominent and includes more target cues.
Furthermore, following~\cite{centernet}, we propose a \textbf{L}iDAR-based 3D Object \textbf{T}racking with \textbf{TR}ansformer framework (LTTR), which is simple but efficient. 
Experiments on KITTI~\cite{KITTI} dataset show that LTTR has outstanding tracking performance and achieves new state-of-the-art performance.

In summary, our contributions are as follows:
\begin{itemize}	
	\item We propose a transformer architecture that explores not only the inter- and intra- relations among different regions within the point cloud but also the relations between different point clouds.
	\item We propose a new 3D object tracking framework based on the transformer architecture, which is simple but efficient.
	\item Extensive experimental results on KITTI dataset show that the proposed method achieves outstanding tracking performances.
\end{itemize}

\section{Related Work}
\label{sec:related}
\subsection{3D Object Tracking}
3D object tracking aims to localize the object in successive frames in 3D space given the initial position. Previous works usually focus on RGB-D data~\cite{Parttracking, rgbdtracking}, which heavily depend on visual features. Recently, with the development of 3D vision methods, there are many LiDAR-based 3D object tracking works~\cite{SC3D, FSiamese, P2B}. For example, Giancola~\etal~\cite{SC3D} used point clouds to track object in LiDAR space based on computing the cosine similarity between template and search branch. However, they ignored the characteristics of the point clouds. Zou~\etal~\cite{FSiamese} leveraged RGB image feature to generate 3D search space, and used point clouds feature to track. Based on~\cite{SC3D}, Qi~\etal~\cite{P2B} proposed a feature fusion module to augment search point features and achieved state-of-the-art tracking performance. In this paper, we explore the inter- and intra- relations among different regions and propagate features between branches to compute region attentions by leveraging the transformer.
\subsection{Vision Transformer}
Due to the great success of Transformer~\cite{Transformer} in natural language processing, recent works start to apply it to vision tasks. Dosovitskiy~\etal~\cite{VIT} proposed ViT to apply a pure transformer in image classification. They split an image into a series of flattened patches and processes the patches by vanilla transformer block to get image cls token. Furthermore, Han~\etal~\cite{TNT} explored the intrinsic structure information inside each patch and achieved higher accuracy than ViT. Chu~\etal~\cite{chu2021conditional} explored the position embedding for ViT and proposed a conditional positional encoding scheme. Liu~\etal~\cite{swin} proposed a shifted windows-based attention and a pure hierarchical backbone which could be used in dense vision tasks. 

Carion~\etal~\cite{DETR} proposed DETR which is the first work to apply the transformer into dense prediction tasks. They applied the transformer architecture into object detection and found the best match between the encoded image embeddings and object queries via the attention module. However, DETR suffers from heavy computation and slow convergence. Zhu~\etal~\cite{deformabledetr} proposed deformable attention to reduce the complexity and speed up convergence, yielding higher performance. There are also other works applying transformers to other tasks, such as visual tracking~\cite{wang2021transformer, chen2021transformer}, multi-object tracking~\cite{transtrack,trackformer}.

\begin{figure}
	\centering
	\includegraphics[width=13cm]{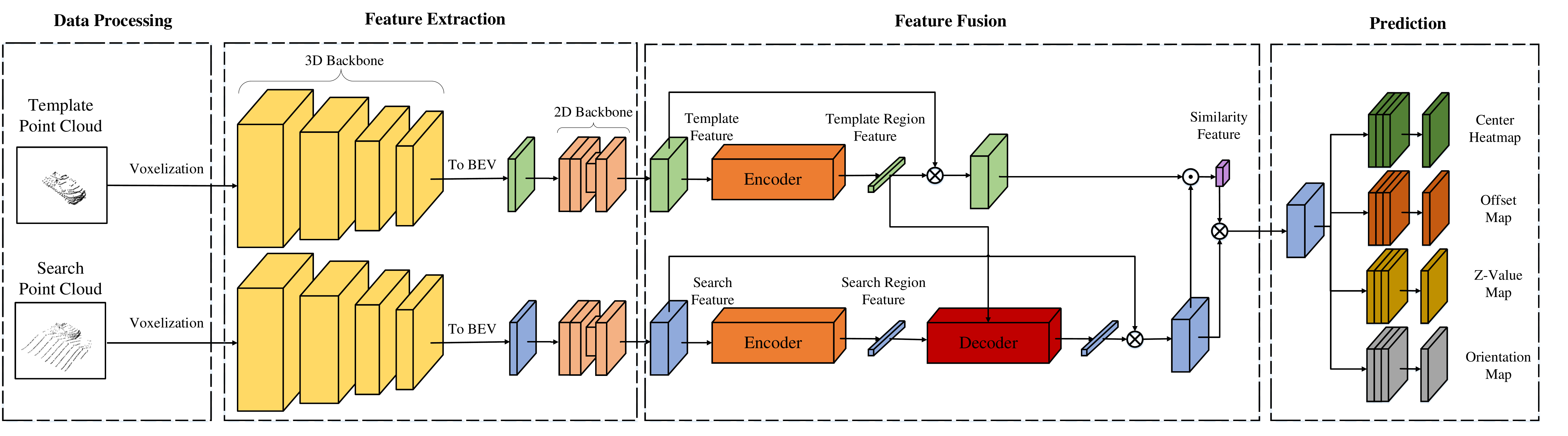}
	\caption{An overview of our LiDAR-based 3D Object Tracking with Transformer framework (LTTR). $\odot$ is the cross-correlation operation, $\otimes$ represents multiplication operation.}\label{fig:network}
	\vspace{-0.1in}
\end{figure}

\section{Method}
\label{sec:method}
In this section, we present the proposed framework, named LTTR. As shown in Figure \ref{fig:network}, the framework consists of data processing, feature extraction, feature fusion, and prediction. We will introduce the details of LTTR in the following subsections.
\subsection{Overall Architecture}
\textbf{Data Processing.}
We adopt the Siamese-like tracking pipeline which inputs template and search point cloud to top and bottom branches respectively. By reading the label,  we obtain the 3D box of the target object and transform the whole scene point clouds into the local coordinate system whose origin is set as the center of the box. After that, we randomly shift the (x,y) of the center of the 3D box to get the training label value in the search branch, then normalize the points into the x-axis of the box in the template branch. Finally, we apply the same 3D range to both branches to get the input pair. The point cloud in the 3D range in the search branch is the search point cloud, and the point cloud in the 3D box in the template branch is the template point cloud. For both branches, we divide the points into regular voxels with a spatial resolution of $W \times L\times H$ and get input $\mathit{I} \in \mathbb{R}^{\mathit{W} \times \mathit{L} \times \mathit{H}}$.

\textbf{Feature Extraction.}
We use the 3D sparse convolution network and 2D convolution network as the backbone network to extract features for both branches. 
Through 3D sparse convolution, the voxels are converted into feature volumes with $8\times$ downsampled sizes. 
By converting the $8\times$ downsampled 3D feature volumes into BEV representation, the final feature map $\mathit{M} \in \mathbb{R}^{\mathit{\frac{W}{8}} \times \mathit{\frac{L}{8}} \times \mathit{F}}$ 
is generated following the 2D backbone network, where $F$ is the feature channels. The weights are sharing between two branches.

\textbf{Feature Fusion.} Subsequently, we update and fuse the search feature $\mathit{M}_{s}$ and the template feature $\mathit{M}_{t}$ in the feature fusion network. As shown in Figure \ref{fig:network}, $\mathit{M}_{s}$ and $\mathit{M}_{t}$ are first fed into the encoder respectively, and then sent into the decoder together. Following \cite{TNT}, the transformer encoder receives $\mathit{M} \in \mathbb{R}^{\mathit{\frac{W}{8}} \times \mathit{\frac{L}{8}} \times \mathit{F}}$ and outputs region feature $\mathit{G} \in \mathbb{R}^{\mathit{N} \times \mathit{D}}$ of channel $D$ with $N$ regions. The transformer decoder propagates information from template regions $\mathit{G}_{t}$ to search regions $\mathit{G}_{s}$ and decodes a fused $\mathit{G}_{s} \in \mathbb{R}^{\mathit{N} \times \mathit{D}}$ through cross-attention. Moreover, we project the region feature $\mathit{G} \in \mathbb{R}^{\mathit{N} \times \mathit{D}}$ to $\mathit{G} \in \mathbb{R}^{\mathit{N} \times \mathit{1}}$ as an attention weight by a fully-connected layer in both branches, and unfold the original feature $\mathit{M}$ to the size of ${\mathit{\frac{W \times L \times F}{64 \times N}} \times \mathit{N}}$ to multiply with $\mathit{G}$. The feature is recovered back to the size of ${\frac{W}{8}} \times \mathit{\frac{L}{8}} \times \mathit{F}$ finally. The details of transformer architecture will be described in Section \ref{sec:Transformer}. Following the depthwise cross-correlation, the similarity feature with size ${\mathit{1} \times \mathit{1} \times \mathit{F}}$ is computed between $\mathit{M}_{s}$ and $\mathit{M}_{t}$. Finally, we multiply the similarity feature with $\mathit{M}_{s}$ to recover feature size for dense prediction.

\textbf{Prediction.} Following~\cite{centernet, afdet}, we use a center-based regression to predict several object properties. The regression consists of four heads, including the center heatmap head, local offset head, z-axis location head, and orientation head. Since our aim is to track the target object, we follow the assumption in~\cite{P2B} that the 3D object size is known.
The heads produce a center heatmap $\mathit{\hat{H}} \in \mathbb{R}^{\mathit{\frac{W}{8}} \times \mathit{\frac{L}{8}} \times \mathit{C}}$, a local offset regression map $\mathit{\hat{O}} \in \mathbb{R}^{\mathit{\frac{W}{8}} \times \mathit{\frac{L}{8}} \times \rm{2}}$, a z-value map $\mathit{\hat{Z}} \in \mathbb{R}^{\mathit{\frac{W}{8}} \times \mathit{\frac{L}{8}} \times \mathit{1}}$ and an orientation map $\hat{\Theta} \in \mathbb{R}^{\mathit{\frac{W}{8}} \times \mathit{\frac{L}{8}} \times \mathit{2}}$ respectively, where $\mathit{C}$ is the number of classes (1 in our tracking task) and orientation includes $sin(\theta)$ and $cos(\theta)$. We follow~\cite{afdet} to set heatmap value for every point $(x,y)$ in the downsampling feature map as:
\begin{equation}
H_{x,y,c} =
\begin{cases}
1, & \text{if $d = 0$}\\
0.8, & \text{if $d = 1$}\\
\frac{1}{d}, & \text{otherwise}
\end{cases}       
\end{equation}
where $d$ is the Euclidean distance calculated between the object center and the point location in the downsample BEV map. A prediction $\mathit{\hat{H}_{x,y,c}=}$ $1$ corresponds to the object center and $\mathit{\hat{H}_{x,y,c}=}$ $0$ corresponds to background. We train the heatmap with focal loss~\cite{focal}:
\begin{equation}\label{equ:2}
\mathcal{L}_{heat} = -\frac{1}{N}\sum_{x,y,c}
\begin{cases}
\left(1 - \hat{H}_{x,y,c}\right)^{\alpha} \log \left(\hat{H}_{x,y,c}\right) \quad\smash{{\text{if ${H}_{x,y,c} = 1$}}} \\
\left(1 - H_{x,y,c}\right)^{\beta}  \left(\hat{H}_{x,y,c} \right)^{\alpha} \quad\quad \ \smash{\raisebox{-1.6ex}{\text{otherwise}}} \\
\quad\quad \log \left( 1 - \hat{H}_{x,y,c}\right), \\
\end{cases}
\end{equation}

For other heads, we use L1 loss:
\begin{equation}
\mathit{\mathcal{L}_{v}} = \frac{1}{N}\sum_{k=1}^{N} \mathit{\left | \hat{V}_{p^{(k)}} - v^{(k)} \right |}.
\end{equation}
where $\mathcal{L}_{v} \in (\mathcal{L}_{off}, \mathcal{L}_{z}, \mathcal{L}_{ori})$, $\hat{V}$ is the true value and $v^{(k)}$ is the predicted value for these heads. Therefore, the overall training loss is
\begin{equation}\label{equ:4}
\mathit{\mathcal{L} = \mathcal{L}_{heat} + \lambda_{off}\mathcal{L}_{off} + \lambda_{z}\mathcal{L}_{z}+ \lambda_{ori}\mathcal{L}_{ori}}
\end{equation}
where $\lambda$ is the regularization parameter for each head.

\subsection{Transformer Architecture}\label{sec:Transformer}
\begin{figure}
	\subfigure[]{
		\includegraphics[width=9cm]{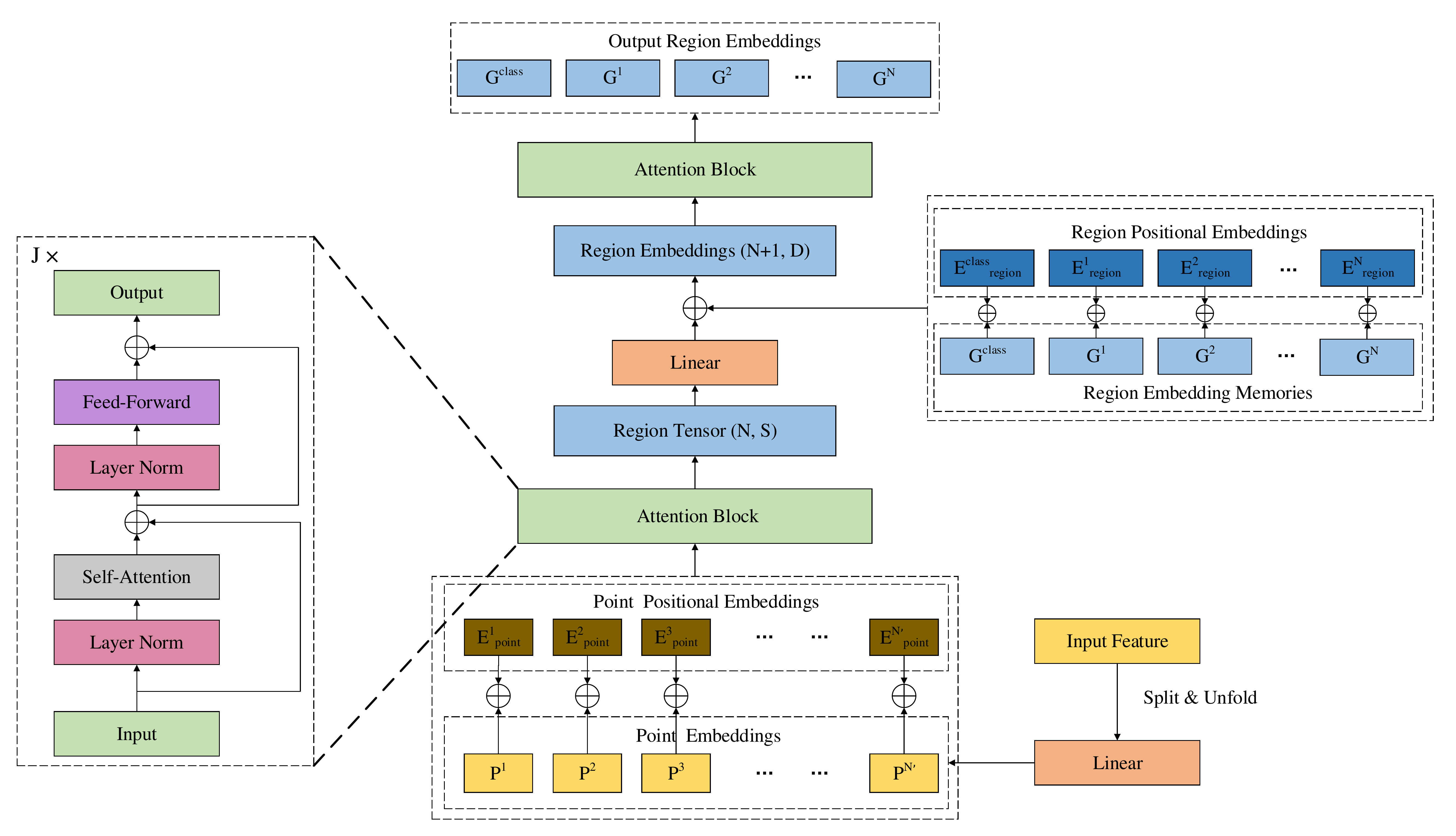}\label{fig:encoder}
	}
	\quad
	\subfigure[]{
		\includegraphics[width=3cm]{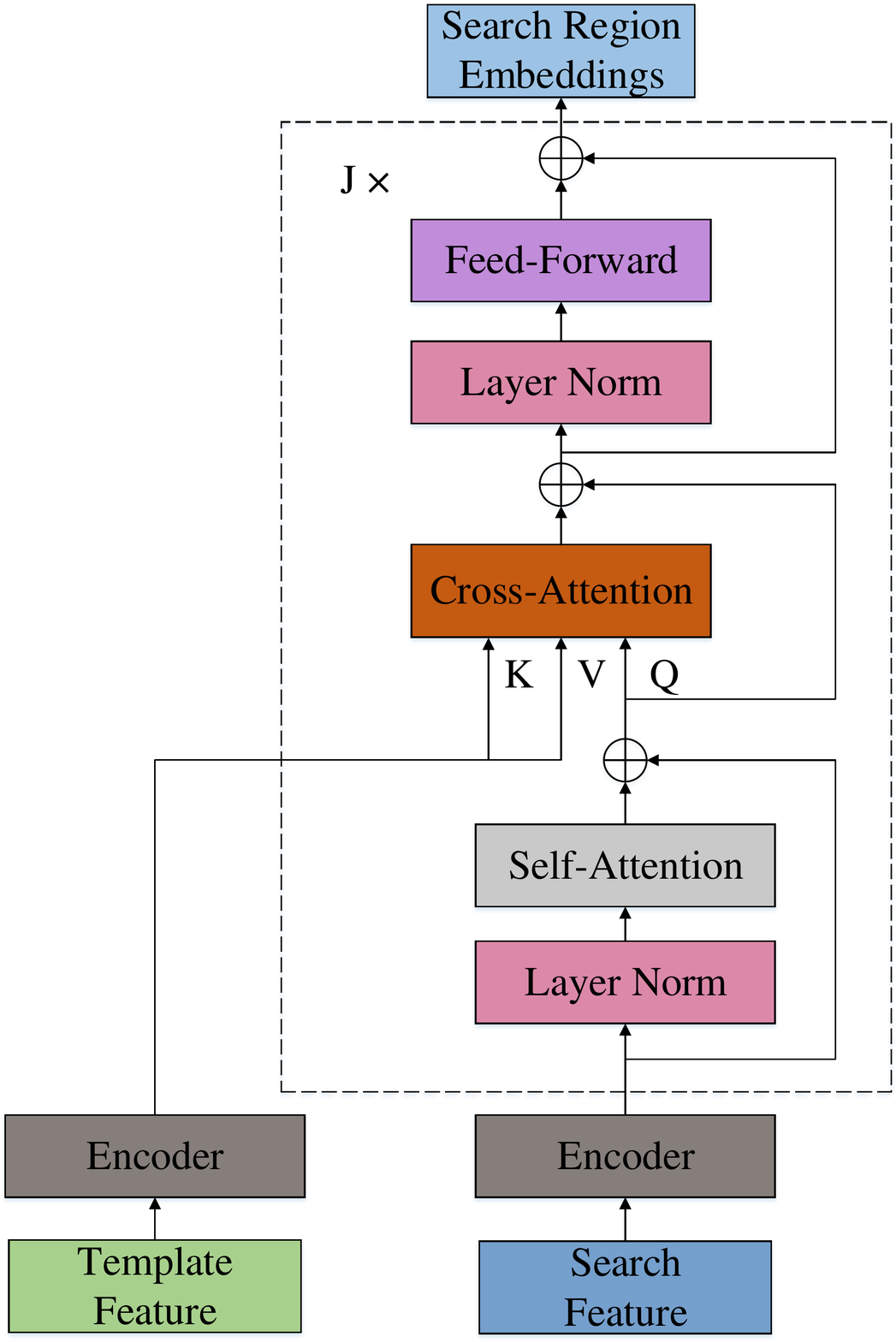}\label{fig:decoder}
	}
	\caption{(a) The transformer encoder. (b)An overview of the proposed transformer architecture.}
	\label{fig:transformer}
	\vspace{-0.1in}
\end{figure}

\textbf{Multi-head Attention.}\label{MHA} Attention function is the core of the transformer, thus we first briefly review the principle of attention. Given query matrix $Q$, key matrix $K$ and value matrix $V$, attention function computes the similarity matrix between query and key, then multiplies value with normalized similarity, defined as:
\begin{equation}
\mathrm{Attention}(Q, K, V) = \mathrm{softmax}(\frac{QK^T}{\sqrt{d_k}})V
\end{equation}
where $d_k$ is the dimension of key.
Meanwhile, multiple heads are usually utilized in the attention function. Multi-head attention (MHA) projects query, key, and value into different feature spaces $h$ times, where $h$ is the number of heads, and computes the attention in parallel for every of these projected queries, keys, and values. The results from different heads are concatenated and projected to the final value. Following \cite{Transformer}, the define of MHA is:
\begin{equation}
\mathrm{MHA}(Q, K, V) = \mathrm{Concat}(\mathrm{head_1}, ..., \mathrm{head_h})W^O
\end{equation}
where $\mathrm{head_i} = \mathrm{Attention}(QW^Q_i, KW^K_i, VW^V_i)$ and $W^O \in \mathbb{R}^{hd_v \times d_{model}}$, $d_v$ is the dimension of value and $d_{model}$ is the dimension of a single head attention.

\textbf{Transformer Encoder.} 
The transformer encoder takes BEV point cloud feature $\mathit{M} \in \mathbb{R}^{\mathit{\frac{W}{8}} \times \mathit{\frac{L}{8}} \times \mathit{F}}$ as its input. Following \cite{TNT}, we first split $M$ into $N$ non-overlapping regions of resolution $(R, R)$ and reshape them to $\mathit{M'} \in \mathbb{R}^{\mathit{N} \times (\mathit{R} \times \mathit{R} \times \mathit{F})}$, where $\mathit{N} = \frac{W}{8R} \times \frac{L}{8R}$. Meanwhile, we also transform each region into the target size $(R', R')$ with point unfold. After applying a linear projection, the sequence of regions can be formed as:
\begin{equation}
\mathcal{U}_0=[U_0^1,U_0^2,\cdots,U_0^{N}]\in\mathbb{R}^{{N}\times (R'\times R'\times S)}
\end{equation}
where $U_0^i\in\mathbb{R}^{R' \times R' \times S}$, $i=1,2,\cdots,N$, and $S$ is the number of channels. Furthermore, each region tensor can also be viewed as a sequence of point tensors:
\begin{equation}
U_0^i=[P_0^{i,1},P_0^{i,2},\cdots,P_0^{i,{N'}}]
\end{equation}
where $N'=R'^{2}$. By utilizing multi-head self-attention, we could explore the intra-relation among regions:
\begin{align}
&\hat{U_j^i} = U_{j-1}^i+\mathrm{MHA}(U_{j-1}^i, U_{j-1}^i, U_{j-1}^i),\\
&U_{j}^i = \hat{U_{j}^i}+\mathrm{FFN}(\hat{U_{j}^i}).
\end{align}
where $j=1,2,\cdots,J$ is the index of the $j$-th layer, $J$ is the total number of layers, and FFN means the feed-forward network, which is a 2-layer MLP module. The point-level MHA builds the local relations among points within one region and produces the region tensor. Additionally, similar to previous vision transformer works~\cite{VIT, TNT, DETR}, we create a set of learnable parameters called region embedding memories $\mathcal{G}_0$ for the region tensors and take them into output as the region representations. Specially, the region embedding memories are added with the region tensors in each layer:
\begin{align}
\mathcal{G}_0&=[G_\text{class},G_0^1,G_0^2,\cdots,G_0^N]\in\mathbb{R}^{(N+1)\times D} \\ 
G_{j-1}^i &= G_{j-1}^i + \mathrm{\Phi}(U_{j-1}^i),
\end{align}
where $G_\text{class}$ is the global point cloud embedding, $G_{j-1}^i\in\mathbb{R}^{D}$, $\Phi$ is the projection function, which is fully-connected layer in our implementation. We random initialize all of the region embedding memories. Meanwhile, we utilize the MHA once again for region embeddings. The mechanism can be summarized as:
\begin{align}
\hat{{G}_j^i} &= {G}_{j-1}^i+\mathrm{MHA}({G}_{j-1}^i, {G}_{j-1}^i, {G}_{l-1}^i),\\
{G}_j^i &= \hat{{G}_{j}^i}+\mathrm{FFN}(\hat{{G}_{j}^i}).
\end{align}
The region-level MHA explores the inter-relation among regions, building the global information of the point cloud. Therefore, the region embedding memories learn the region representation by adding to region tensors and being sent into the MHA during training. Meanwhile, although $G_\text{class}$ does not have a corresponding region tensor to add, it can also capture the global information by exchanging information with the other region embeddings through the region-level MHA.
Furthermore, we use standard learnable 1D position embeddings to add to embeddings as follows:
\begin{equation}
T = T + E
\end{equation}
where $T \in (G, U)$, $E \in (E_{region}, E_{point})$, $E_{region}\in\mathbb{R}^{(N+1) \times D}$ and  $E_{point}\in\mathbb{R}^{N' \times S}$. Both the region and point position embeddings are added to the corresponding embeddings before MHA and are shared across the same data level, thus the local and global spatial information can be maintained. The whole process is shown in Figure \ref{fig:encoder}.

By processing both points and regions, the encoder explores the local information across points within regions and global relations across regions, producing $\mathit{G} \in \mathbb{R}^{(\mathit{N}+1) \times \mathit{D}}$for each point cloud feature $\mathit{M}$. We take $
G=[G_0^1,G_0^2,\cdots,G_0^{N}]\in\mathbb{R}^{N\times D}$ as the input of the decoder.

\textbf{Transformer Decoder.}
The above encoder processes template and search features separately, thus the information only flows within the point cloud itself. To build the inter-relation between point clouds and exchange information across branches, we further utilize a transformer decoder to fuse features. The decoder fuses features by propagating template region feature $\mathit{G}_{t}$ to search region feature $\mathit{G}_{s}$. 
The decoder first updates the search region feature $\mathit{G}_{s}$ by self-attention mechanism, then computes the similarity among regions from search and template point clouds based on the cross-attention mechanism. Specially, the decoder takes $\mathit{G}_{s}$ as the query and $\mathit{G}_{t}$ as key and value through the cross-attention, the fused search region feature $\mathit{G}_{s}$ is generated following a feed-forward layer. The decoder is shown in Figure \ref{fig:decoder} and can be summarized as:
\begin{align}
	&\hat{G_{s}} = G_{s}+\mathrm{MHA}(G_{s}, G_{s}, G_{s}),\\	
	&\tilde{G_{s}} = \hat{G_{s}}+\mathrm{MHA}(\hat{G_{s}}, G_{t}, G_{t}),\\
	&G_{s} = \tilde{G_{s}}+\mathrm{FFN}(\tilde{G_{s}}).
\end{align}
Through the decoder, the search and template region features exchange region information, which makes the search region feature include much more information of the target object and computes the region attention. To have clear representations, the layer norm operation is not represented in the above equations.

\section{Experiments}
\label{sec:exper}
\subsection{Datasets and Evaluation}
We use KITTI tracking dataset~\cite{KITTI} as the benchmark and follow \cite{P2B} in data split. We also use One Pass Evaluation (OPE) as evaluation metric, including Success and Precision.
\subsection{Implementation Details}
In data processing, we set point cloud range as [-3.2m, 3.2m], [-3.2m, 3.2m], [-3m, 1m] along x, y, z axis, and set voxel size as [0.025m, 0.025m, 0.05m]. The template and search points are voxelized following \cite{VoxelNet}. A maximum of five points are randomly sampled from each voxel. Meanwhile, we use the same backbone as ~\cite{VoxelNet, second}. In regression, each head consists of four convolution layers to predict and the heatmap head is followed by a sigmoid function to generate the final score. Following the training setting of the popular codebase OpenPCDet~\cite{openpcdet}, we train the network end-to-end with 80 epochs and 36 batch. In loss setting, we set $\alpha = 2$, $\beta = 4$ in Equation \ref{equ:2}, and set $\lambda_{z}=1.5$, $\lambda_{off} = \lambda_{ori} = 1$ in Equation \ref{equ:4}.

\begin{table}[t]
	\begin{adjustbox}{center}
		\begin{tabular}{ccccccc}\toprule
			Method   & Reference  & LiDAR & RGB & Success & Precision & FPS  \\ \hline
			SC3D~\cite{SC3D}& CVPR2019 & $\surd$      &     & 41.3    & 57.9    & 1.8 \\
			F-Siamese~\cite{FSiamese}  & IROS2020  & $\surd$   & $\surd$ & 37.1    & 50.6    & -\\
			P2B~\cite{P2B}  & CVPR2020  & $\surd$      &     & 56.2    & 72.8     & \textbf{45.5} \\ \hline
			LTTR(Ours)  & - & $\surd$    &  & \textbf{65.0}   & \textbf{77.1}     & 22.6\\ \hline
		\end{tabular}
	\end{adjustbox}
	\vspace{-0.1in}
	\caption{Comprehensive comparison with state-of-the-art trackers on Car category.}
	\label{table:Car}
	\vspace{-0.2in}
\end{table}

\begin{table}[t]
	\begin{adjustbox}{center}
		\begin{tabular}{ccccccc}\toprule
			%\hline
			& Method     & Car  & Pedestrian & Van & Cyclist & Mean \\ 
			& Frame Number     & 6424  & 6088 & 1248 & 308 & 14068 \\ \hline
			\multirow{4}{*}{Success}   
			& SC3D~\cite{SC3D} & 41.3 &  18.2 & 40.4    & 41.5   &  31.2    \\
			& F-Siamese~\cite{FSiamese}  & 37.1 & 16.2           & -    & 47.0        & -    \\
			
			& P2B~\cite{P2B}   & 56.2 & 28.7   & \textbf{40.8}    & 32.1  & 42.4     \\
			& LTTR(Ours) & \textbf{65.0} &   \textbf{33.2} &  35.8   & \textbf{66.2}   &   \textbf{48.7}   \\ \hline
			\multirow{4}{*}{Precision}  
			& SC3D~\cite{SC3D}     & 57.9 & 37.8     & 47.0    & 70.4        & 48.5     \\
			& F-Siamese~\cite{FSiamese}  & 50.6 &  32.2          & -    &  77.2       &  -   \\
			& P2B~\cite{P2B}       & 72.8 & 49.6     & \textbf{48.4}   & 44.7        & 60.0     \\
			& LTTR(Ours) & \textbf{77.1} &\textbf{56.8}    & 45.6    &   \textbf{89.9}      & \textbf{65.8}     \\ \hline
		\end{tabular}
	\end{adjustbox}
	\vspace{-0.1in}
	\caption{Extensive comparisons with state-of-the-art trackers on multiple categories.}
	\label{table:allkitti}
	\vspace{-0.1in}
\end{table}

\begin{figure}[t]
	\centering
	\includegraphics[width=7cm]{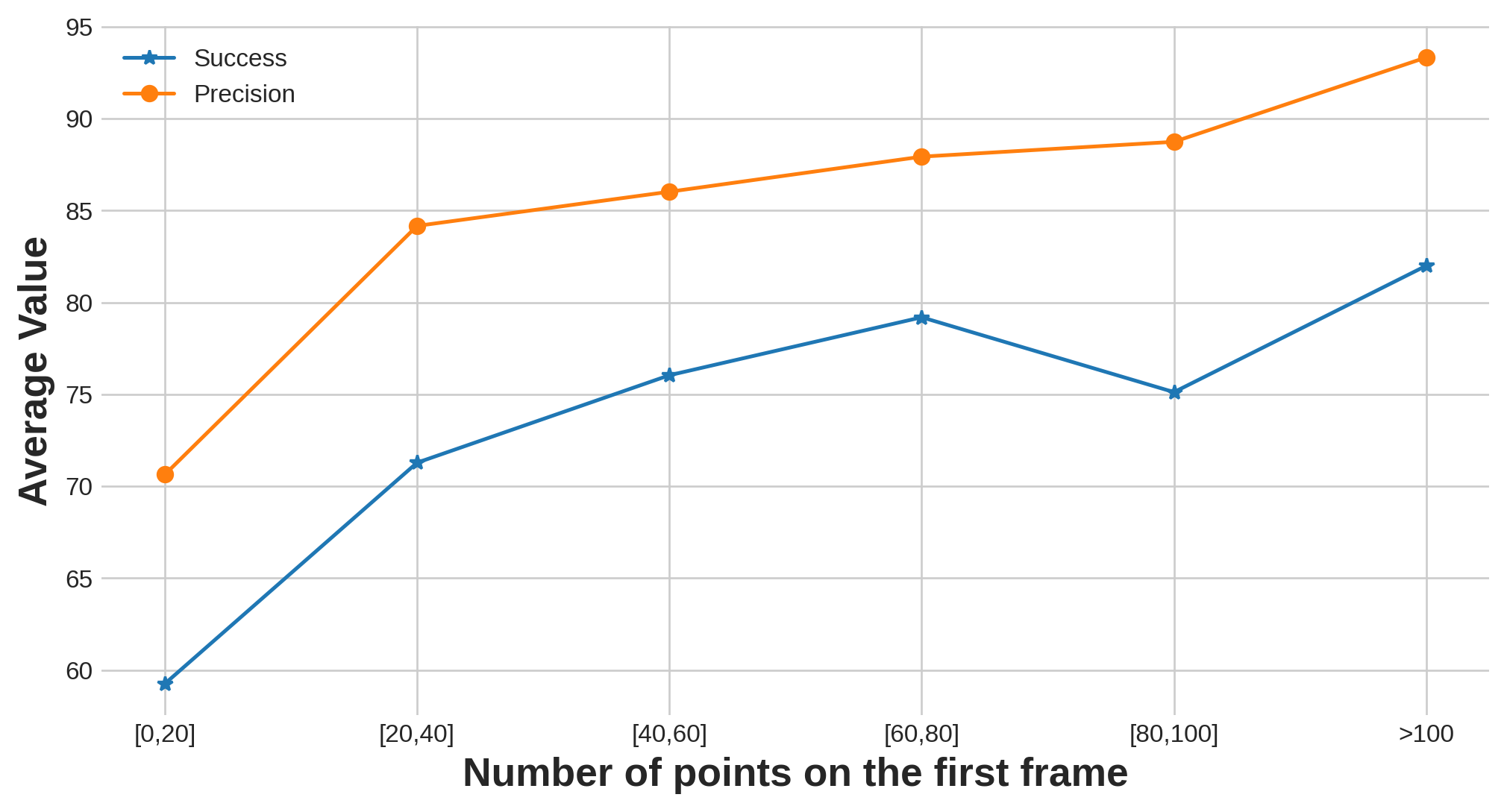}
	\caption{The influence of the number of points on the first frame car.}\label{fig:first_points}
	\vspace{-0.2in}
\end{figure}

\subsection{State-of-the-art Comparisons}
We compare our LTTR with previous state-of-the-art methods on KITTI dataset. As shown in Table \ref{table:Car}, our approach surpasses the previous methods by +8.8\% Success and +4.3\% Precision respectively in the Car category. Additionally, LTTR achieves a real-time running speed. We also report multiple categories tracking results on KITTI dataset, including Pedestrian, Van, and Cyclist. As shown in Table \ref{table:allkitti}, our method outperforms P2B~\cite{P2B} by 5\% on average. In particular, LTTR shows its advantages on objects with a small size, e.g. Pedestrian and Cyclist, surpassing previous methods by a large margin. Considering the difference among these categories, our method is a general and efficient method for different categories.

We also report the influence of the number of the first frame point in the Car category. As shown in Figure \ref{fig:first_points}, with more points, LTTR has a higher performance. We believe that more points in the first frame give the network enough information about the target to track.

\begin{wraptable}[7]{r}{6cm}
	\vspace{-2mm}
	\scriptsize
	\centering
	\begin{tabular}{cccll}
		\cline{1-3}
		Different Network Version & Success       & Precision     &  &  \\ \cline{1-3}
		Baseline                      & 57.2          & 70.9          &  &  \\
		Encoder (w/o Decoder)    & ${\text{60.6}}_{3.4\%\uparrow} $         & ${\text{71.9}}_{1.0\%\uparrow} $          &  &  \\
		Encoder +  Decoder (Max)     & ${\text{64.2}}_{7.0\%\uparrow} $          & $\textbf{77.3}_{6.4\%\uparrow} $ &  &  \\
		Encoder +  Decoder            & $\textbf{65.0}_{7.8\%\uparrow} $ & $\text{77.1}_{6.2\%\uparrow} $          &  &  \\ \cline{1-3}
	\end{tabular}
	\vspace{-0.1in}
	\caption{Ablative study of our transformer architecture.}	
	\label{table:componet}
\end{wraptable}

\subsection{Ablation Study}
In this section, we ablate the proposed method on the Car category of KITTI dataset. We first ablate the transformer network to verify the influence of the encoder and decoder. We introduce a baseline version and a Max-Decoder version. The baseline version does not have any transformer component, and the Max-Decoder version inputs $\mathit{G}_{class}$ instead of $G=[G_0^1,G_0^2,\cdots,G_0^{N}]$ in the template branch to the decoder. Moreover, we compare the different numbers of heads, layers and region sizes in the transformer to validate our design choices. Finally, we compare different backbones and regression heads to explore their influence on the proposed method.

\textbf{Effect of Encoder.}
As shown in Table \ref{table:componet}, with the transformer encoder, the performance has +3.4\% and +1.0\% gains on Success and Precision respectively. The result indicates the effectiveness of our encoder to build the local and global relations of point cloud. It is worth noting that even without any transformer component, our baseline version has a competitive performance with the state-of-the-art methods. 

\textbf{Effect of Decoder.}
We further add the transformer decoder. Specially, we evaluate two decoder versions different in template input. As shown in Table \ref{table:componet}, both of them bring significant performance improvements. However, compared to Max-version, our version has a balanced result between Success and Precision. We believe that the Max-version loses the inter-relation among regions of the point cloud due to its single global input. 

\begin{wraptable}[14]{r}{6cm}
	\vspace{-0.4mm}
	\scriptsize
	\centering
	\begin{tabular}{cccc}\toprule
		&   & Success & Precision \\ \hline
		\multirow{5}{*}{Head Number}  & 1 & 61.0    & 73.7      \\
		& 2 & 61.6    & 74.6      \\
		& 4 & 62.1      & 75.3      \\
		& 8 & \textbf{65.0}    & 77.1      \\
		& 12 & 63.8    & \textbf{78.3}      \\ \hline
		\multirow{5}{*}{Layer Number} & 1 & \textbf{65.0}    & \textbf{77.1}      \\
		& 2 & 61.9    & 74.3      \\
		& 4 & 62.7    & 75.7      \\
		& 6 & 61.1    & 73.8      \\
		& 8 & 60.2    & 73.2      \\ \hline
		\multirow{3}{*}{Region Size}  & 1 & 60.5    & 73.3      \\
		& 4 & 63.6    & 76.7      \\
		& 16 & \textbf{65.0}    & \textbf{77.1}      \\ \hline
	\end{tabular}
	\vspace{0.1in}
	\caption{Ablative study of our transformer architecture.}	
	\label{table:tarchitecture}
\end{wraptable}

\textbf{Structure Modifications.}
We also discuss the details of our transformer structure as shown in Table \ref{table:tarchitecture}, including the number of heads, number of encoder/decoder layers and the region size. All experimental networks have a complete encoder and decoder component. 
For the number of heads, we observe that heads=8 achieves the best performance, while increase heads to 12 results in a decrease in Success but an increase in Precision. The results indicate that MHA is efficient in our transformer architecture, as discussed in Section \ref{MHA}, but too many heads may result in degeneration in orientation prediction. 
Meanwhile, stacking more layers does not bring in performance improvement but has more parameters and lower speed. We speculate that more layers may divide the template and search features into different feature subspaces. Different from detection task, the tracking task has two input branches and tracks the object based on their similarity. Therefore, tracking method requires the template and search features to be in the same feature spaces to have a better similarity computation.
Additionally, with a small region size, the performance of the network degenerates to the encoder version. We believe that the smaller size generates more regions and leads to the decoder not being able to exchange global information effectively. Thus, we use the max non-overlapping size $R=16$ for higher performance.

\textbf{Backbone.}
We also make modifications to the backbone to explore whether the performance could be further improved by increasing the parameters in the backbone. Our backbone follows Second~\cite{second}, a backbone baseline in 3D vision for its simple architecture and wide use~\cite{pv_rcnn, VoxelNet, fastpoint}. The baseline includes 3D and 2D backbones to process voxels and BEV features respectively. The 3D backbone is termed as BaseVoxel and the 2D backbone is termed as BaseBEV. In this comparison experiment, we use the resnet-manner version of BaseVoxel in OpenPCDet~\cite{openpcdet} and termed it as ResVoxel, which adds a residual path in every sparse block of BaseVoxel. Meanwhile, we add one convolution block to BaseBEV and term it as DeepBEV. Therefore, the ResVoxel has more parameters than BaseVoxel, and DeepBEV is deeper than BaseBEV. However, as Table \ref{table:backbone} shows, with the network going deeper and the total parameters becoming larger, the performance does not have improved but decreased. We speculate that more parameters in the backbone may hinder the transformer to capture useful information, thus our baseline backbone could achieve better performance with fewer parameters comparing to these modifications.

\begin{table}[t]
	\centering
	\scriptsize	
	\begin{tabular}{cccccc}\toprule
		3D Backbone & 2D Backbone& 3D Params & 2D Params & Success & Precision  \\ \hline
		\multirow{2}{*}{BaseVoxel}  
		& BaseBEV  & 1.280K  &    8.266M&  \textbf{65.0} &   \textbf{77.1}       \\
		& DeepBEV & 1.280K & 12.988M &    62.4     &  76.4                  \\ \hline
		\multirow{2}{*}{ResVoxel} 
		& BaseBEV   &   2.656K & 8.266M  &   62.6      &     75.4              \\
		& DeepBEV   &   2.656K & 12.988M  &  60.2       &   73.8                \\ \hline
	\end{tabular}
	\vspace{0.05in}
	\caption{Ablative study of different 3D and 2D backbones.}
	\label{table:backbone}
	%\vspace{-0.2in}
\end{table}

\textbf{Regression Head.}
	We compare our center-based regression head with an anchor-based counterpart. For anchor-based regression, we follow the setting of Second~\cite{second}. Specially, for every location, we set two anchors with 0 degrees and 90 degrees, and the thresholds for 
	\begin{wraptable}[6]{t}{6cm}
		\setlength{\tabcolsep}{1mm}
		%\vspace{-0.1mm}
		\scriptsize
		\centering
		\begin{tabular}{cccccc}\toprule
			&   & Success & Precision  \\ \hline
			& Center-based & 65.0    & \textbf{77.1}   \\
			& Anchor-based & \textbf{65.3}    & 75.1   \\ \hline
		\end{tabular}
		\vspace{0.2in}
		\caption{Comparison of different regression heads.}	
		\label{table:regression_head}
	\end{wraptable}
	positive and negative are 0.6 and 0.45 respectively. As shown in Table \ref{table:regression_head}, the anchor-based regression improves the Success 0.3 points while reduces the Precision 2.0 points. The result shows that the center-based regression has better prediction results in location but is weaker in rotation regression compared to anchor-based regression. We believe this is because of the pre-defined anchor rotation degree in anchor-based regression. Meanwhile, the anchor-based regression needs more hyper-parameters in the anchor setting which needs fine-tuning. Therefore, to have fewer hyper-parameters and more balanced tracking results, we adopt the center-based regression.

\begin{figure}[t]
	\centering
	\includegraphics[width=9cm]{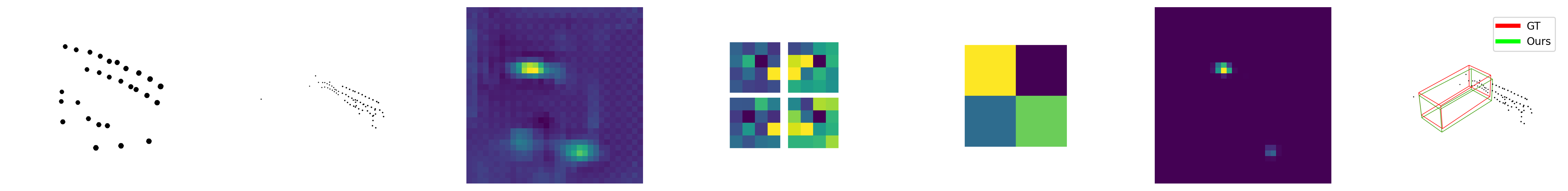}
	\includegraphics[width=9cm]{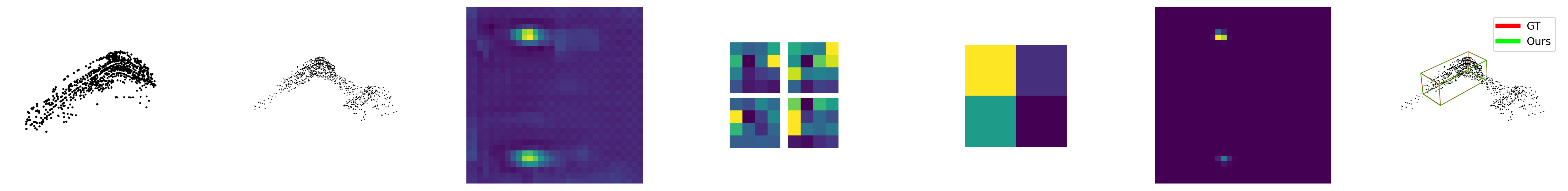}
	\includegraphics[width=9cm]{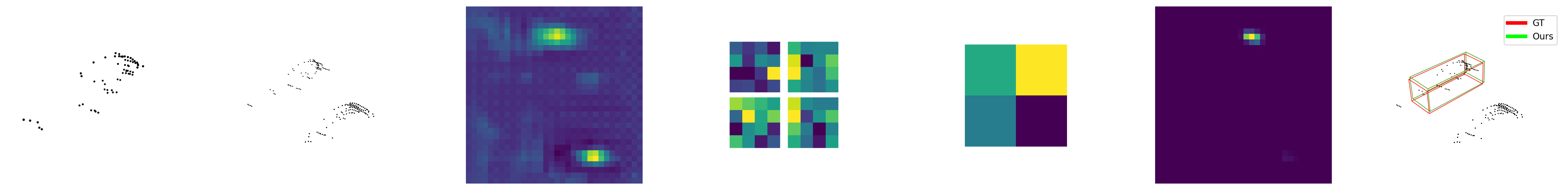}
	\caption{Visualization results. There are template point cloud, search point cloud, heatmap without transformer, point-level attention, region-level attention, heatmap with transformer and the predicted boxes from left to right.}\label{fig:heatmap}
	\vspace{-0.2in}
\end{figure}

\subsection{Qualitative Visualization}
To explore the effect of the transformer network, we visualized the predicted heatmap and the transformer point and region weights, as shown in Figure \ref{fig:heatmap}. Compared to the heatmap without transformer, the heatmap with transformer accurately finds the target position with the help of point and region attentions, avoiding the false track. The results verify the effectiveness of the proposed transformer network, even for target object with sparse points.

\section{Conclusions}
In this paper, we present LTTR, a novel tracking framework based on the transformer. Through the transformer network, LTTR builds local information and global relation within the point cloud, explores the inter-relation between point clouds, and predicts the 3D bounding box of the target object by a center-based regression. Comprehensive experiments on KITTI dataset demonstrate that our method achieves new state-of-the-art performance. In the future, we will investigate how to integrate temporal information into our method.

\textbf{Acknowledgments}
This work was supported by National Natural Science Foundation of China (62073066, U20A20197), Science and Technology on Near-Surface Detection Laboratory (6142414200208), the Fundamental Research Funds for the Central Universities (N182608003), Major Special Science and Technology Project of Liaoning Province (No.2019JH1/10100026), and Aeronautical Science Foundation of China (No.201941050001).
\bibliography{egbib}
\end{document}